\def\year{2019}\relax
\documentclass[letterpaper]{article} 
\usepackage{aaai19}  
\usepackage{times}  
\usepackage{helvet}  
\usepackage{courier}  
\usepackage{url}  
\usepackage{graphicx}  

\graphicspath{ {images/} }
\usepackage{amsmath}
\usepackage{amssymb}
\usepackage{multirow}
\usepackage{subcaption}

\title{An Affect-Rich Neural Conversational Model with Biased Attention \\
and Weighted Cross-Entropy Loss}
\author{Peixiang Zhong,$^{1, 2}$ Di Wang,$^1$ Chunyan Miao$^{1,2,3}$\\
$^1$Joint NTU-UBC Research Centre of Excellence in Active Living for the Elderly\\
$^2$Alibaba-NTU Singapore Joint Research Institute\\
$^3$School of Computer Science and Engineering\\
Nanyang Technological University, Singapore\\
peixiang001@e.ntu.edu.sg, \{wangdi, ascymiao\}@ntu.edu.sg
}
\begin{document}
\maketitle
\begin{abstract}
Affect conveys important implicit information in human communication. Having the capability to correctly express affect during human-machine conversations is one of the major milestones in artificial intelligence. In recent years, extensive research on open-domain neural conversational models has been conducted. However, embedding affect into such models is still under explored. In this paper, we propose an end-to-end affect-rich open-domain neural conversational model that produces responses not only appropriate in syntax and semantics, but also with rich affect. Our model extends the Seq2Seq model and adopts VAD (Valence, Arousal and Dominance) affective notations to embed each word with affects. In addition, our model considers the effect of negators and intensifiers via a novel affective attention mechanism, which biases attention towards affect-rich words in input sentences. Lastly, we train our model with an affect-incorporated objective function to encourage the generation of affect-rich words in the output responses. Evaluations based on both perplexity and human evaluations show that our model outperforms the state-of-the-art baseline model of comparable size in producing natural and affect-rich responses.
\end{abstract}
\section{Introduction}
\label{sec:introduction}

Affect is a psychological experience of feeling or emotion. As a vital part of human intelligence, having the capability to recognize, understand and express affect and emotions like human has been arguably one of the major milestones in artificial intelligence \cite{picard1997affective}.

Open-domain conversational models aim to generate coherent and meaningful responses when given user input sentences. In recent years, neural network based generative conversational models relying on Sequence-to-Sequence network (Seq2Seq) \cite{sutskever2014sequence} have been widely adopted due to its success in neural machine translation. Seq2Seq based conversational models have the advantages of end-to-end training paradigm and unrestricted response space over conventional retrieval-based models. To make neural conversational models more engaging, various techniques have been proposed, such as using stochastic latent variable \cite{serban2017hierarchical} to promote response diversity and encoding topic \cite{xing2017topic} into conversational models to produce more coherent responses.

However, embedding affect into neural conversational models has been seldom explored, despite that it has many benefits such as improving user satisfaction \cite{callejas2011predicting}, fewer breakdowns \cite{martinovski2003breakdown}, and more engaged conversations \cite{robison2009evaluating}. For real-world applications, \citeauthor{fitzpatrick2017delivering} (\citeyear{fitzpatrick2017delivering}) developed a rule-based empathic chatbot to deliver cognitive behavior therapy to young adults with depression and anxiety, and obtained significant results on depression reduction. Despite of these benefits, there are a few challenges in the affect embedding in neural conversational models that existing approaches fail to address: (\romannumeral 1) It is difficult to capture the emotion of a sentence, partly because negators and intensifiers often change its polarity and strength. Handling negators and intensifiers properly still remains as a challenge in sentiment analysis. (\romannumeral 2) It is difficult to embed emotions naturally in responses with correct grammar and semantics \cite{ghosh2017affect}.

\begin{figure*}[!t]
    \centering
    \begin{subfigure}[b]{0.3\textwidth}
        \centering
        \includegraphics[scale=0.28]{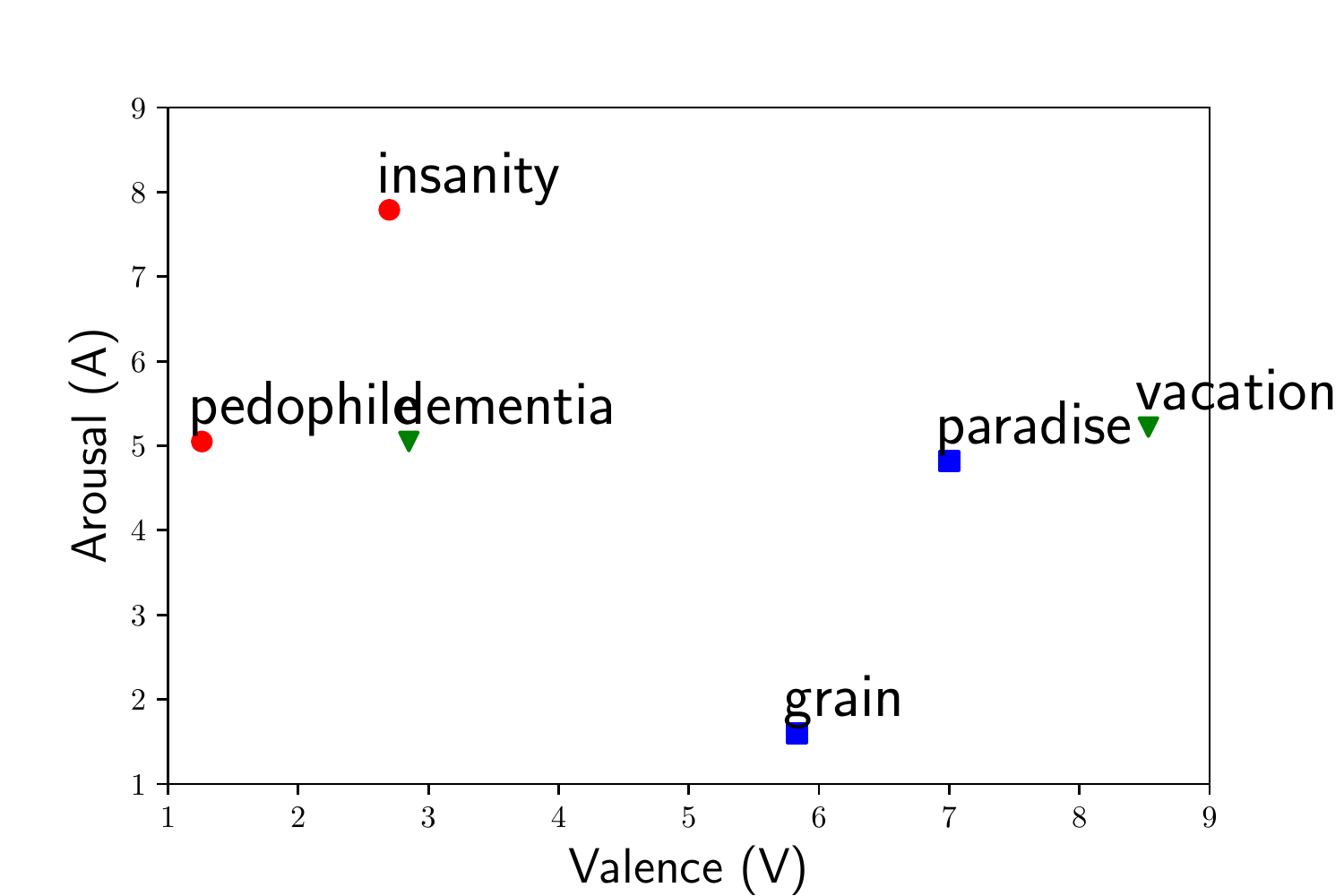}
        \caption{V-A ratings}
    \end{subfigure}
    \begin{subfigure}[b]{0.3\textwidth}
        \centering
        \includegraphics[scale=0.28]{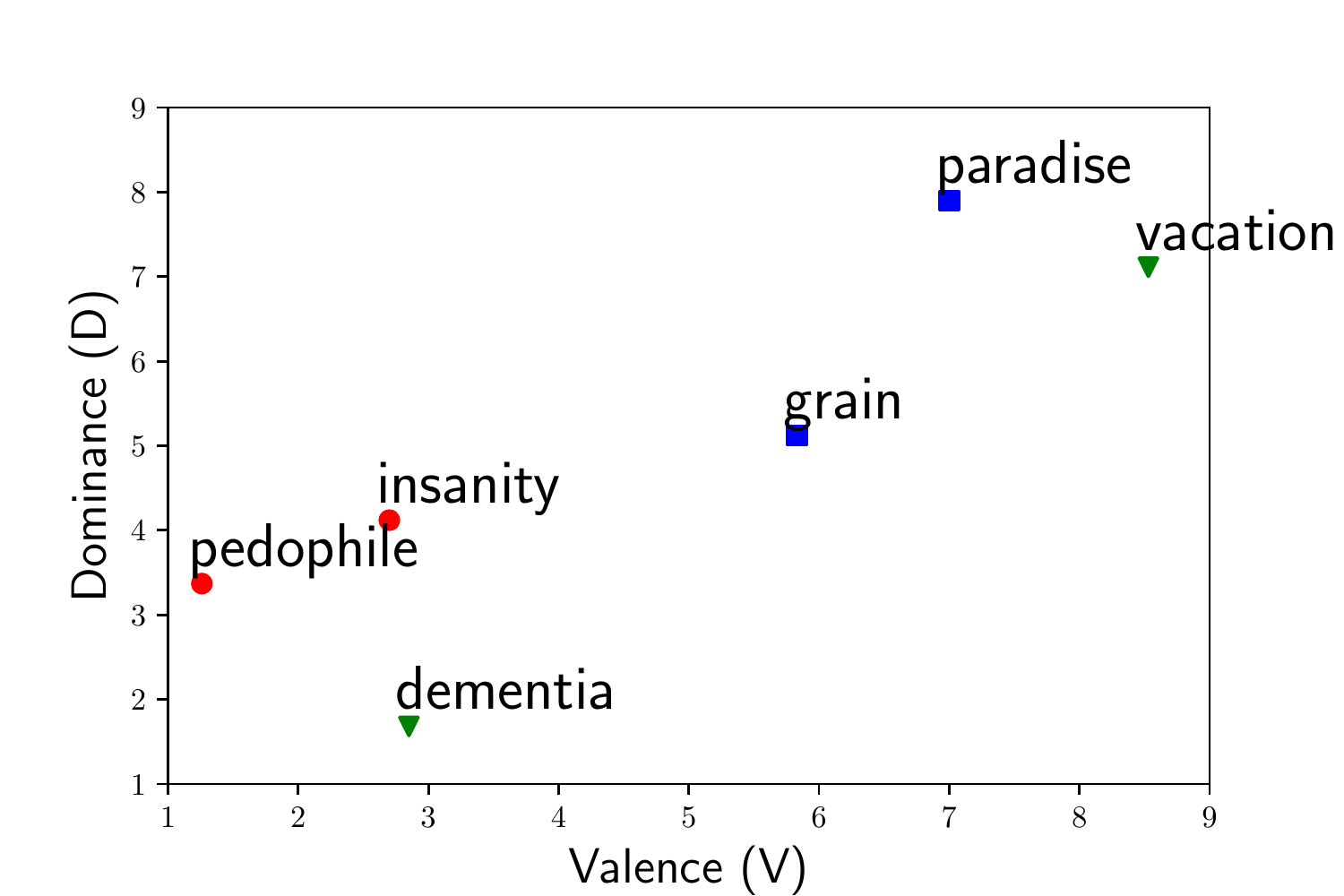}
        \caption{V-D ratings}
    \end{subfigure}
    \begin{subfigure}[b]{0.3\textwidth}
        \centering
        \includegraphics[scale=0.28]{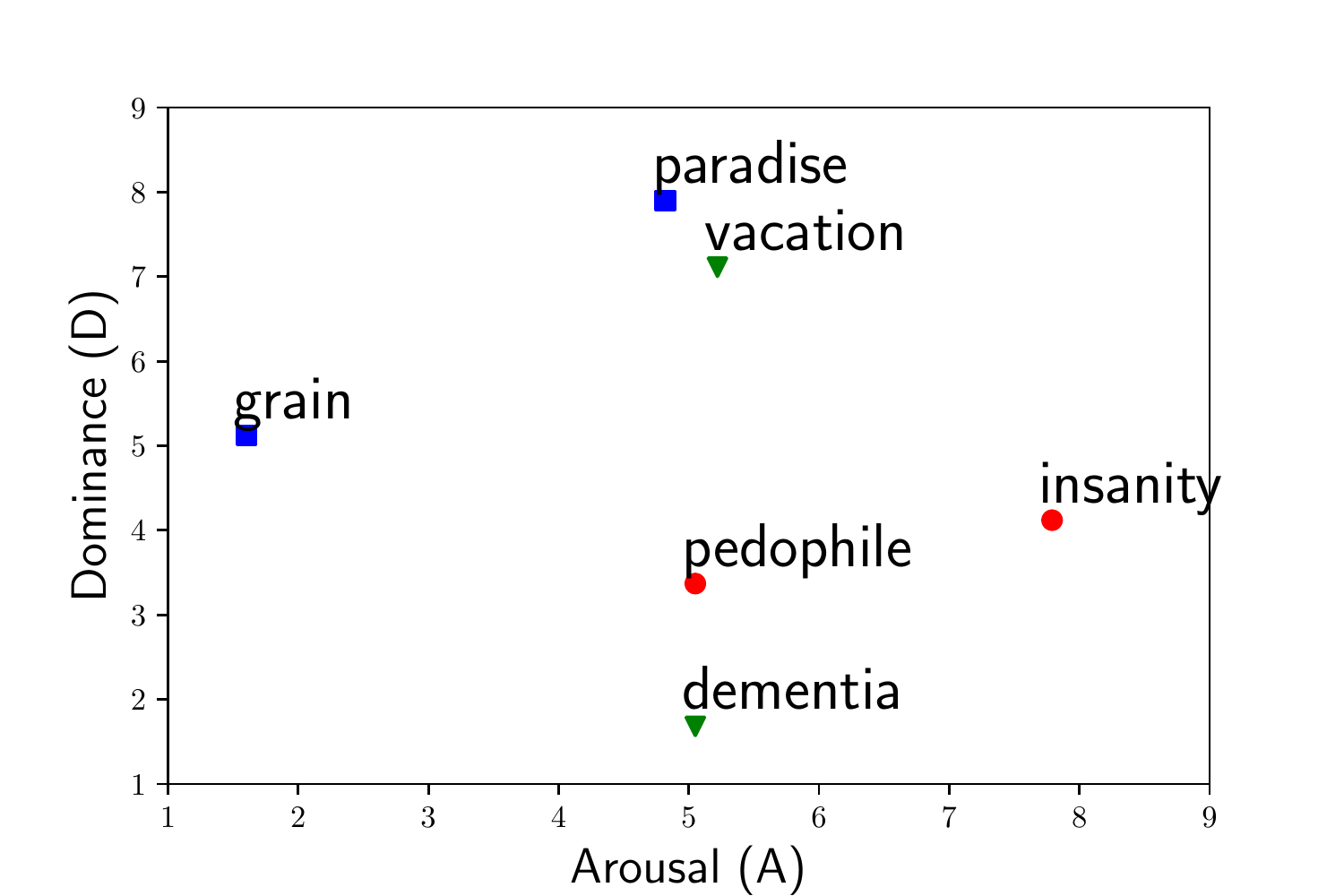}
        \caption{A-D ratings}
    \end{subfigure}
    \caption{2D plot of words with either highest or lowest ratings in valence (V), arousal (A) or dominance (D) in the corpus.}
    \label{fig: VAD plot}
\end{figure*}

In this paper, we propose an end-to-end single-turn open-domain neural conversational model to address the aforementioned challenges to produce responses that are natural and affect-rich. Our model extends Seq2Seq model with attention \cite{luong2015effective}. We leverage an external corpus \cite{warriner2013norms} to provide affect knowledge for each word in the Valence, Arousal and Dominance (VAD) dimensions \cite{mehrabian1996pleasure}. We then incorporate the affect knowledge into the embedding layer of our model. VAD notation has been widely used as a dimensional representation of human emotions in psychology and various computational models, e.g., \cite{wang2016modeling,tang2017eeg}. 2D plots of selected words with extreme VAD values are shown in Figure \ref{fig: VAD plot}. To capture the effect of negators and intensifiers, we propose a novel biased attention mechanism that explicitly considers negators and intensifiers in attention computation. To maintain correct grammar and semantics, we train our Seq2Seq model with a weighted cross-entropy loss that encourages the generation of affect-rich words without degrading language fluency.

Our main contributions are summarized as follows:

\begin{itemize}
  \item For the first time, we propose a novel affective attention mechanism to incorporate the effect of negators and intensifiers in conversation modeling. Our mechanism introduces only a small number of additional parameters.
  \item For the first time, we apply weighted cross-entropy loss in conversation modeling. Our affect-incorporated weights achieve a good balance between language fluency and emotion quality in model responses. Our empirical study does not show performance degradation in language fluency while producing affect-rich words.
  \item Overall, we propose \textit{Affect-Rich Seq2Seq} (AR-S2S), a novel end-to-end affect-rich open-domain neural conversational model incorporating external affect knowledge. Human preference test shows that our model is preferred over the state-of-the-art baseline model in terms of both content quality and emotion quality by a large margin.
\end{itemize}
\section{Related Work}
\label{sec:related work}
Prior studies on affective conversational systems mainly focused on rule-based systems, which require an extensive hand-crafted rule base. For example, \citeauthor{ochs2008empathic} (\citeyear{ochs2008empathic}) designed an empathetic virtual agent that can express emotions based on cognitive appraisal theories \cite{hewstone2001introduction}, which require numerous event-handling rules to be implemented. Another example is the Affect Listeners \cite{skowron2010affect}, which are conversational systems aiming to detect and adapt to the affective states of users. However, their detection and adaptation mechanisms heavily rely on hand-crafted features such as letter capitalization, punctuation and emoticons. 

In recent years, there is an emerging research trend in end-to-end neural network based generative conversational systems \cite{vinyals2015neural,shang2015neural}. To improve the content quality of neural conversational models, many techniques have been proposed, such as improving response diversity using Conditional Variational Autoencoders (CVAE) \cite{learning2017zhao} and encoding commonsense knowledge using external facts corpus \cite{knowledge2018Ghazvininejad}.

However, few work investigated the problems in improving the emotion quality of neural conversational models. Emotional Chatting Machine (ECM) \cite{zhou2018emotional} is a Seq2Seq conversational model that generates responses with user-input emotions. It employs an internal memory module to model implicit emotional changes and an external memory module to help generate more explicit emotional words. The main objective of ECM is to produce responses according to explicit user-input emotions. While our model focuses on enriching affect in generated responses. 
Similar to ECM, Mojitalk \cite{zhou2018mojitalk} presents a few generative models, including Seq2Seq, CVAE and Reinforced CVAE, to generate responses according to explicit user-input emojis. Both ECM and Mojitalk do not consider emotions in input sentences when generating emotional responses. In comparison, our model considers them naturally with focuses on affect-rich words and avoids an additional step of determining which emotion to respond with during conversations. 
\citeauthor{asghar2017affective} (\citeyear{asghar2017affective}) introduces a Seq2Seq model with three extensions to incorporate affects into conversations. Similar to their work, we also adopt the approach of using VAD embedding to encode affects. However, we perform extra preprocessing on VAD embedding to improve model performance. In addition, we specifically consider the effect of negators and intensifiers via a novel affective attention mechanism when generating affect-rich responses.
\begin{figure*}[!t]
\centering
\includegraphics[scale=0.58]{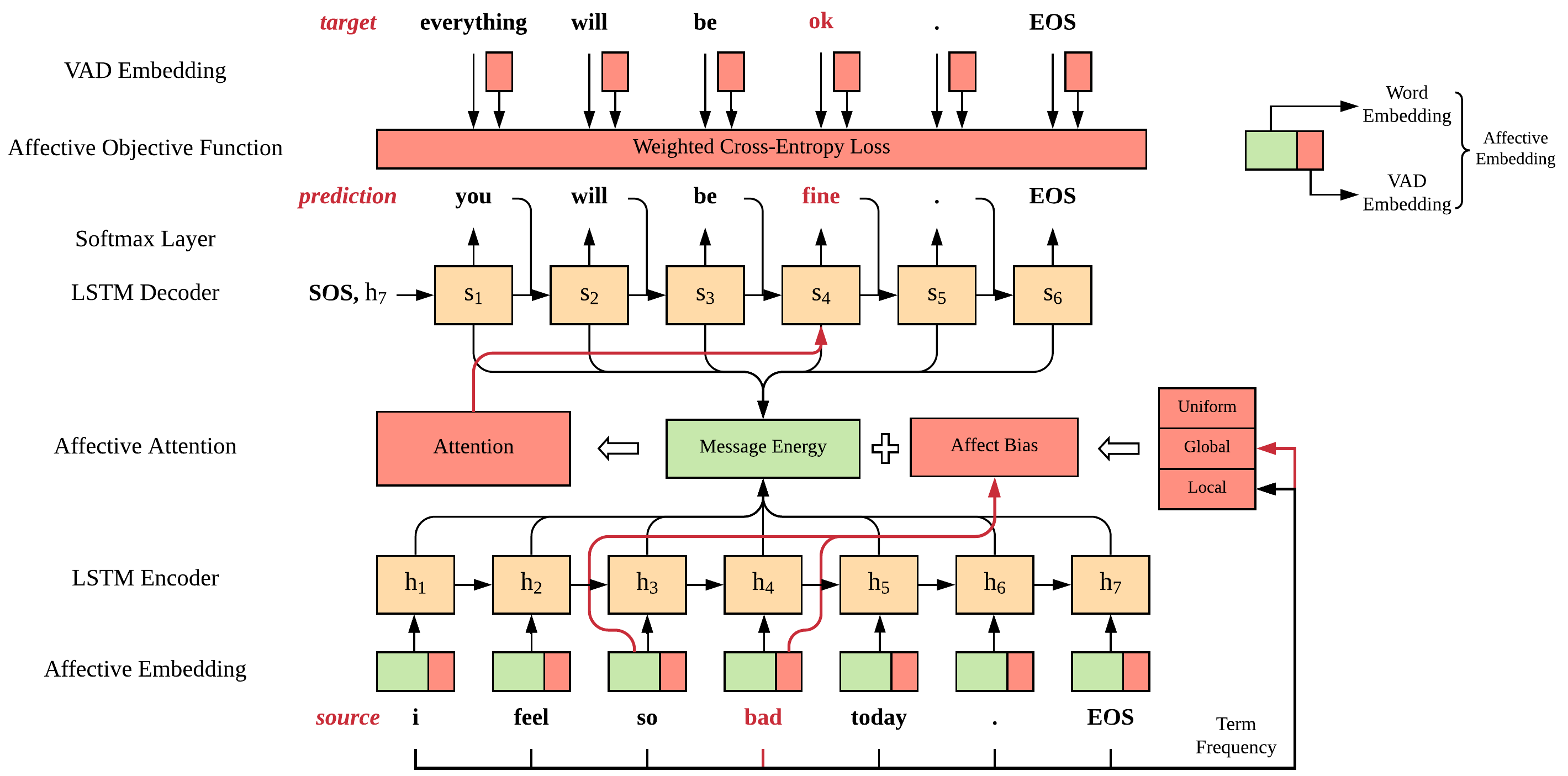}
\caption{Overall architecture of our proposed AR-S2S. This diagram illustrates decoding ``fine'' and affect bias for ``bad''.}
\label{fig: architecture}
\end{figure*}
\section{Seq2Seq with Attention}
\label{sec:seq2seq}

Prior to introducing our proposed model, we briefly describe the vanilla Seq2Seq model with attention. Seq2Seq model is a neural network model mapping the input sequence to the output sequence. Specifically, it uses a Recurrent Neural Network (RNN) encoder to encode the variable length input sequence $X = (x_1, x_2, ..., x_T)$ as a vector of fixed dimensionality $\mathbf{h_T}$ and an RNN decoder to decode $\mathbf{h_T}$ as the variable length output sequence $Y = (y_1, y_2, ..., y_{T^{'}})$. The objective function of Seq2Seq is to maximize
\begin{equation}
\label{eqn: objective}
\begin{split}
p(Y|X) &= p(y_1|\mathbf{h}_T) \prod_{{t^{'}}=2}^{T^{'}}p(y_{t^{'}}|\mathbf{h}_T,y_1, ..., y_{t^{'}-1}), \\
\mathbf{h}_t &= f(\mathbf{h}_{t-1}, x_t), \forall t  = 1, 2, ..., T,
\end{split}
\end{equation}
where $\mathbf{h_t}$ denotes the hidden state of input sequence at time step $t$ and $\mathbf{h_0}$ is usually initialized as a zero vector. Function $f$ denotes a non-linear transformation, which usually takes the form of recurrent models such as Long Short-Term Memory (LSTM) \cite{hochreiter1997long} or Gated Recurrent Units (GRU) \cite{cho-al-emnlp14}.

After encoding $X$ as $\mathbf{h_T}$, the decoder updates its decoder hidden state $\mathbf{s_{t^{'}}}$ by taking the previous hidden state $\mathbf{s}_{{t^{'}}-1}$ and previous output $y_{{t^{'}}-1}$ as inputs: 
\begin{equation}
\mathbf{s}_{{t^{'}}} = g(\mathbf{s}_{{t^{'}}-1}, y_{{t^{'}}-1}), \forall {t^{'}}  = 1, 2, ..., {T^{'}},
\end{equation}
where $g$ is another recurrent model, $s_0 = \mathbf{h_T}$, and $y_0$ is the start of sequence (SOS) token.

The output word probability in equation (\ref{eqn: objective}) is given by 
\begin{equation}
\label{eqn: word probability}
p(y_{t^{'}}) = \textit{softmax}(\mathbf{W}^o\mathbf{s}_{t^{'}}), \forall {t^{'}}  = 1, 2, ..., {T^{'}},
\end{equation}
where $\mathbf{W}^o$ denotes a model parameter.

The attention mechanism \cite{luong2015effective} is proposed to solve the problem of limited representation power of the final input hidden state $\mathbf{h}_T$ on which the entire decoding process is conditioned. Specifically, the attention mechanism focuses on different parts of the input sequence by computing a context vector $\mathbf{c}_{t^{'}}$ at each decoding time step $t^{'}, \forall t^{'} = 1, 2, ..., T^{'}$, as the weighted average of all input hidden states $\mathbf{h_t}, \forall t = 1, 2, ..., T$, as follows:
\begin{equation}
\label{eqn: vanilla attention}
\mathbf{c}_{t^{'}} = \sum_{t=1}^{T}\alpha_{t^{'}t}\mathbf{h}_t,
\end{equation}
where the alignment vector $\alpha_{t^{'}t}$ is given by
\begin{equation}
\label{eqn: alignment}
\alpha_{t^{'}t} = \frac{\exp(e_{t^{'}t})}{\sum_{k=1}^{T} \exp(e_{t^{'}k})},
\end{equation}
where $e_{t^{'}t} = \textit{score}(\mathbf{h}_t, \mathbf{s}_{t^{'}})$ is the message energy function that computes the energy or score between input hidden state $\mathbf{h}_t$ and output hidden state $\mathbf{s}_{t^{'}}$. This message energy function is usually implemented as a Multilayer Perceptron (MLP). In our case, we use a simple dot product operation due to its fast training and good performance \cite{luong2015effective}.

The context vector $\mathbf{c}_{t^{'}}$ is then concatenated with the decoder hidden state $\mathbf{s}_{t^{'}}$ to form an attentional hidden state $\mathbf{\hat{s}_{t^{'}}}$ as follows:
\begin{equation}
\mathbf{\hat{s}_{t^{'}}} = \tanh(\mathbf{W}^c[\mathbf{c}_{t^{'}};\mathbf{s}_{t^{'}}]),
\end{equation}
where $[;]$ denotes vector concatenation. Finally, $\mathbf{\hat{s}_{t^{'}}}$ replaces $\mathbf{s}_{t^{'}}$ in equation (\ref{eqn: word probability}) to compute the output word probability.
\section{Affect-Rich Seq2Seq Model}
\label{sec:affective seq2seq}
In this section, we present our proposed model to produce affect-rich responses, which falls outside the capability of vanilla Seq2Seq models. The overall model architecture is illustrated in Figure \ref{fig: architecture}.
\subsection{Affective Embedding}
\label{sec:affective embedding}
\begin{table}[!t]
\small
\centering
\begin{tabular}{p{1.5cm}|p{0.8cm}|p{4cm}}
 \hline
 \textbf{Dimensions} & \textbf{Values} & \textbf{Interpretations}\\
 \hline
 Valence &3 - 7&pleasant - unpleasant\\
 Arousal &3 - 7&low intensity - high intensity\\
 Dominance &3 - 7&submissive - dominant\\
 \hline
\end{tabular}
\caption{Interpretations of clipped VAD embeddings.}
\label{table: vad}
\end{table}

Our model adopts Valence, Arousal and Dominance (VAD) \cite{mehrabian1996pleasure} embedding to encode word affects as vectors of size $3$ from an annotated lemma-VAD pairs corpus \cite{warriner2013norms}. This corpus comprises 13,915 lemmas with VAD values annotated in the $[1, 9]$ scale. To leverage this corpus, we assign VAD values to words based on their lemmas. To increase coverage, we extend the corpus to 23,825 lemmas by assigning the average VAD values of their synonyms to absent lemmas. Furthermore, we empirically clip VAD values of all words to the $[3, 7]$ interval to prevent words with extreme VAD values from repeatedly showing in the generated responses, as observed in our preliminary experiments. The interpretations of clipped VAD embedding are presented in Table \ref{table: vad}. For example, word ``nice'' is associated with the clipped VAD values: (V: 6.95, A: 3.53, D: 6.47). For words whose lemmas are not in the extended corpus, comprising approximately 10\% of the entire training vocabulary, we assign them VAD values of $[5, 3, 5]$, which are the clipped values of a neutral word. Note that a value of $3$ in arousal (A) dimension is regarded as neutral because it has zero emotional intensity.

Finally, to remove bias, we normalize VAD embedding as $\overline{\textit{VAD}}(x_t) = \textit{VAD}(x_t) - [5, 3, 5]$, where $\textit{VAD}(x_t) \in \mathbb{R}^{3}$ is the VAD embedding of word $x_t$. We incorporate VAD embedding by concatenation as follows: 
\begin{equation}
\mathbf{e}(x_t) = [\mathbf{x_t}; \lambda \overline{\textit{VAD}}(x_t)],
\end{equation}
where $\mathbf{x_t} \in \mathbb{R}^m$ denotes the word embedding of $x_t$, $\mathbf{e}(x_t) \in \mathbb{R}^{m+3}$ denotes the final affective embedding of $x_t$, $m$ denotes the dimensionality of word vectors, and $\lambda \in \mathbb{R_+}$ denotes the affect embedding strength hyper-parameter to tune the strength of VAD embeddings.

It is worth noting that the lemmas in our corpus were selected across multiple domains and are quite neutral \cite{brysbaert2009moving}. In addition, languages other than English, such as Spanish, Dutch, Finish, etc., also have such lemma-VAD pairs corpus, although in smaller sizes. Hence, our proposed conversational model has great potential to be directly applied to other languages.
\subsection{Affective Attention}
\label{sec:affective attention}
To incorporate affect into attention naturally, we make the intuitive assumption that humans pay extra attention on affect-rich words during conversations. Specifically, our model biases attention towards affect-rich words in the input sentences, as well as considers the effect of negators and intensifiers. Our model employs an affect bias $\eta$ augmenting the affective strength of each word in the input sentences into the energy function (see equation (\ref{eqn: alignment})) as follows:
\begin{equation}
\label{eqn: energy function}
e_{t^{'}t} = {\mathbf{h}_t}^T \mathbf{s}_{t^{'}} + \eta_{t},
\end{equation}
where ${\mathbf{h}_t}^T \mathbf{s}_{t^{'}}$ denotes the conventional dot product energy function and $\eta_{t}$ is defined as
\begin{equation}
\label{eqn: affective attention}
\begin{split}
\eta_{t}&=\gamma||\mu(x_t)(1 + \beta) \otimes \overline{\textit{VAD}}(x_t)||_2^2, \\
\beta&=\tanh(\mathbf{W}^b\mathbf{x_{t-1}}),
\end{split}
\end{equation}
where $\otimes$ denotes element-wise multiplication, $||.||_k$ denotes $l_k$ norm, $\mathbf{W}^{b} \in \mathbb{R}^{3 \times m}$ denotes a model parameter, $\beta \in \mathbb{R}^{3}$ is a scaling factor in V, A and D dimensions in the $[-1, 1]$ interval to scale the normalized VAD values of the current input word, $\gamma \in \mathbb{R}_{+}$ denotes the affective attention coefficient controlling the magnitude of affect bias towards affect-rich words in the input sentence, and $\mu(x_t) \in \mathbb{R}$ in the $[0, 1]$ interval denotes a measure of term importance of $x_t$ (see the following paragraph).

\subsubsection{Term Importance}

The introduction of term importance $\mu(x_t)$ as weights in computing affective attention is inspired by the sentence embedding work \cite{arora2016simple}, where a simple weighted sum of word embedding algorithm with weights being smoothed inverse term frequency can achieve good performance in textual similarity tasks. Term frequency has been widely adopted in information retrieval to compute the importance of a word. In our model, we propose three approaches, namely ``uniform importance" (ui), ``global importance" (gi), and ``local importance" (li) to compute $\mu(x_t)$:
\begin{equation}
  \mu(x_t) =
  \begin{cases}
    1 & \text{ui} \\
    a/(a + p(x_t)) & \text{gi} \\
    \frac{\log(1/(p(x_t) + \epsilon))}{\sum_{t=1}^{t=T}{\log(1/(p(x_t) + \epsilon))}} & \text{li}
  \end{cases},
\label{eqn: term importance}
\end{equation}
where $p(x_t)$ denotes the term frequency of $x_t$ in the training corpus, $a$ denotes a smoothing constant that is usually set to $10^{-3}$ as suggested by \citeauthor{arora2016simple} (\citeyear{arora2016simple}), and $\epsilon$ is another small smoothing constant with value $10^{-8}$. We take the log function in $\mu_{li}(x_t)$ to prevent rare words from dominating the importance.  

\subsubsection{Modeling Negators and Intensifiers}

\begin{figure}[!t]
\centering
\hspace{-0.3cm}
\includegraphics[scale=0.28]{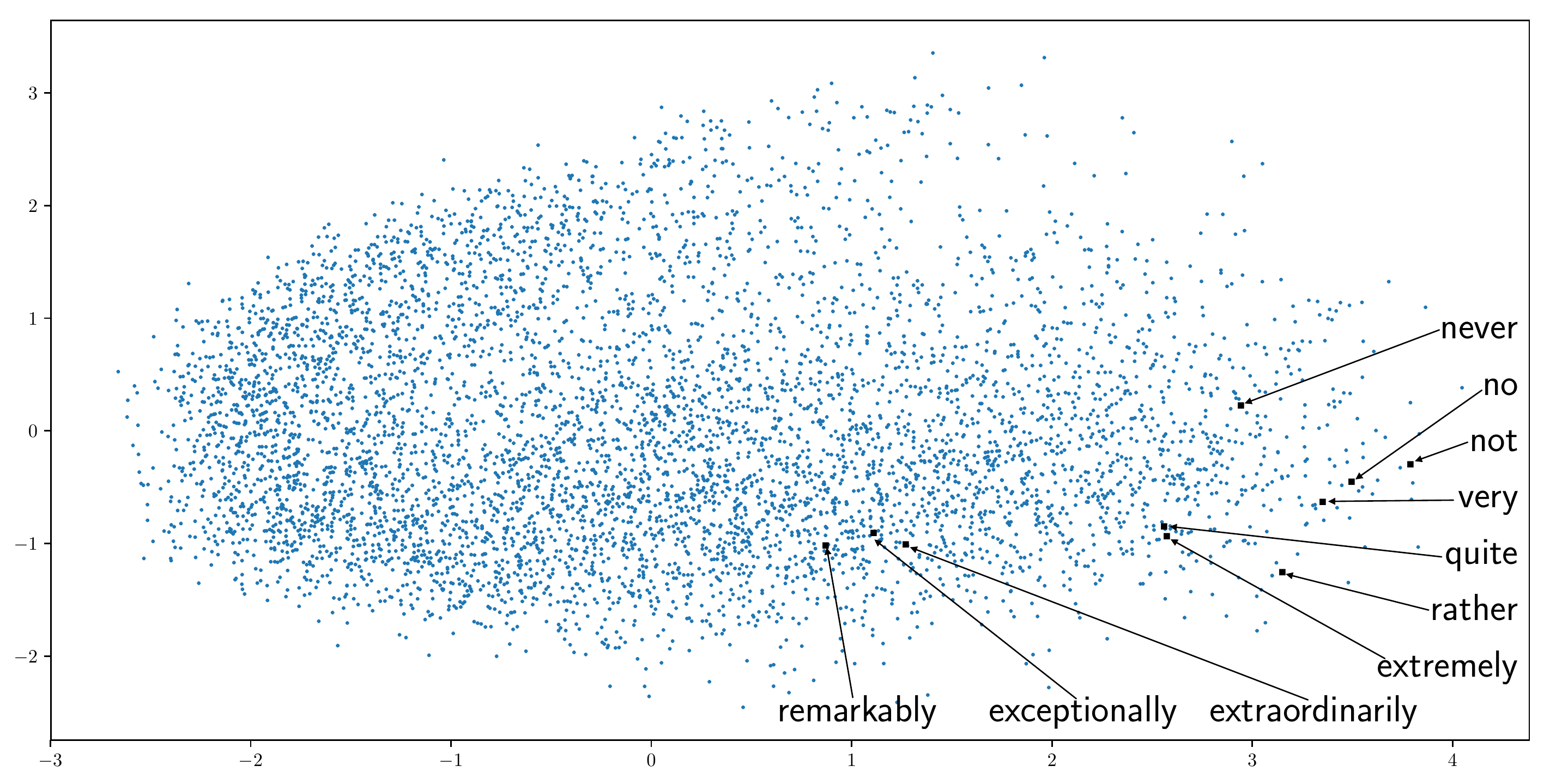}
\caption{2D plot of the most frequent 30,000 words in our training dataset in GloVe embedding after PCA. Selected common negators and intensifiers are annotated in text.}
\label{fig: intensifier}
\end{figure}

The introduction of $\beta$ in equation (\ref{eqn: affective attention}) is to model the affect changes caused by negators and intensifiers. Often, negators make positive words negative but with much lower intensity, and make negative words less negative \cite{kiritchenko2016the}. Thus, $\beta$ is expected to be negative for negators because negators tend to reduce the affect intensity of the following word (e.g., ``not bad"). Intensifiers usually adjust the intensities of positive words and negative words but do not flip their polarities \cite{carrillo2013emotion}. As a result, $\beta$ for extreme intensifiers (e.g., ``extremely") is expected to be larger than $\beta$ for less extreme intensifiers (e.g., ``very"). To specifically consider these phenomena, $\beta$ is modeled to be a nonlinear transformation through the word vector of $x_{t-1}$. This idea is inspired by the observation that common negators and intensifiers share some common underlying properties in their word vector representations. Figure \ref{fig: intensifier} shows that several common negators and intensifiers tend to cluster together in 2D plots in GloVe embedding \cite{pennington2014glove} after applying Principle Component Analysis (PCA). 

Note that our affective attention only considers unigram negators and intensifiers, however, they are empirically found as the majority of all negators and intensifiers. Statistics based on our training set indicate that the unigram intensifier ``very'' occurs 364,913 times, in comparison, the composite intensifier ``not very'' only occurs 2,838 times.
\subsection{Affective Objective Function}
\label{sec:affective objective function}
The conventional objective function of seq2seq model is to maximize the probability of target response Y given input sequence X measured by cross-entropy loss. To encourage the generation of affect-rich words, we incorporate VAD embedding of words into cross-entropy loss as follows:

\begin{equation}
\label{eqn: affective objective 1}
\Psi_{t^{'}} = -|V|\frac{1+\delta ||\overline{\textit{VAD}}(y_{t^{'}})||_2}{\sum_{\hat{y}_{t^{'}} \in V} (1+\delta ||\overline{\textit{VAD}}(\hat{y}_{t^{'}})||_2)}\log(p(y_{t^{'}})),
\end{equation}
where $t^{'} = 1, 2, ..., T^{'}$, $\Psi_{t^{'}}$ denotes the affective loss at decoding time step $t^{'}$, $y_{t^{'}}$ denotes the target token at decoding time step $t^{'}$,  $V$ denotes the dataset vocabulary, and $\delta$ denotes a hyper-parameter named affective loss coefficient, which regulates the contribution of VAD embedding. 

Our proposed affective loss is essentially a weighted cross-entropy loss. The weights are constant and positively correlated with VAD strengths in $l_2$ norm. The weight normalization is applied to ensure that our weights do not alter the overall learning rate during optimization. Intuitively, our affective loss encourages affect-rich words to obtain higher output probability, which effectively introduces a probability bias into the decoder language model towards affect-rich words. This bias is controlled by our affective loss coefficient $\delta$. When $\delta=0$, our affective loss falls back to the conventional unweighted cross-entropy loss. 

It is worth noting that our weighted cross-entropy loss incorporating external word knowledge, i.e., VAD in our case, is simple but effective in controlling the response style. Our loss function has many other potential application areas such as controlled neural text generation.
\section{Experimental Evaluation}
\label{sec:experiment}
In this section, we present our datasets, evaluation methods, experimental results and discussions. Following the experimental setup presented in \cite{zhou2018emotional}, we conduct \textbf{model component test (MCT)} to examine the effectiveness of our proposed affective attention and affective objective function in generating affect-rich responses. In addition, we conduct \textbf{preference test (PT)} between our best model (\textbf{AR-S2S}) and the state-of-the-art baseline of comparable model size to compare model responses. Finally, we conduct \textbf{sensitivity analysis} on the hyper-parameters introduced in our model to analyze their impacts on language fluency and the number of distinct affect-rich words produced. 
\subsection{Datasets}
\label{sec:dataset}
We use OpenSubtitles dataset \cite{tiedemann2009news} as our training dataset due to its large size. We use relatively less noisy Cornell Movie Dialog Corpus dataset \cite{cristian2011chameleons} as our validation dataset for more reliable tuning. We use DailyDialog dataset \cite{li2017dailydialog} for testing to examine model generalizations in different corpus domains. 

The pairs in the training dataset are selected by a simple rule that the input sentence ends with a question mark and the time interval between the pair of input and output sentences is less than 20 seconds. In addition, sound sequences such as ``\textit{BANG}" are removed. These pairs are then expanded (e.g., isn't $\,\to\,$ is not), tokenized, and special symbols and numbers were removed. Finally, the pairs with either input or output sentence longer than 20 words are removed. The validation and testing datasets are preprocessed by word expansion, tokenization and removal of special symbols and numbers. Since we are modeling single-turn dialogue system, only the first two utterances from each dialogue session in the testing dataset are extracted because using utterances in the middle would require context to respond.

After data preprocessing, we randomly select 5 million pairs from OpenSubtitles dataset as the training dataset with a vocabulary comprising the most 30,000 frequent words, covering 98.89\% of all tokens. We randomly sample 100K pairs from Cornell Movie Dialog Corpus dataset for validation and 10K pairs from DailyDialog dataset for testing.
\subsection{Evaluation Methods}
\label{sec:evaluation}
We adopt perplexity metric to measure the language fluency of a conversational model, as it is the only well-established automatic evaluation method in conversation modeling. Other metrics such as BLEU \cite{papineni2002bleu} do not correlate well with human judgments \cite{liu2016how}. A model with lower perplexity indicates that it is more confident about the generated responses. Note that a model with low perplexity does not guarantee to be a good conversational model because it may achieve so by always generating short responses.

To qualitatively examine model performance, we conduct widely adopted human evaluations. We randomly sample 100 input sentences from the testing dataset. For each input sentence, we then randomize the order of the responses generated by each comparison model. For each response, five human annotators are asked to evaluate two aspects: 
\begin{itemize}
\item \textbf{+2}: (content) The response has correct grammar and is relevant and natural / (emotion) The response has adequate and appropriate emotions conveyed.
\item \textbf{+1}: (content) The response has correct grammar but is too universal / (emotion) The response has inadequate but appropriate emotions conveyed.
\item \textbf{0}: (content) The response has either grammar errors or is completely irrelevant / (emotion) The response has either no or inappropriate emotions conveyed.
\end{itemize}
\subsection{Experiment 1: Model Component Test (MCT)}
\label{sec:MCT}
We compare the following models to examine the performance of our proposed affective attention and affective objective function on model perplexity and human evaluations:

\textbf{S2S}: The standard Seq2Seq model with attention.

\textbf{S2S-UI}, \textbf{S2S-GI}, \textbf{S2S-LI}: The standard Seq2Seq model with our proposed affective attention using $\mu_{ui}$, $\mu_{gi}$ and $\mu_{li}$ (see equation (\ref{eqn: term importance})), respectively.

\textbf{S2S-AO}: The standard Seq2Seq model with attention and our proposed affective objective function (see equation (\ref{eqn: affective objective 1})).

\textbf{AR-S2S}: our best model, which incorporates both $\mu_{li}$ and affective objective function.

All models have a word embedding of size 1027 (1024 + 3) and hidden size of 1024. Both encoder and decoder have two layers of bi-directional LSTM. All models implement affective embedding. Parameters $\lambda$, $\delta$ and $a$ are set to 0.1, 0.15 and $10^{-3}$, respectively. Parameter $\gamma$ for S2S-UI, S2S-GI and S2S-LI are set to 0.5, 1 and 5, respectively. The beam size is set to 20. Note that all models implement the maximum mutual information (MMI) objective function \cite{li2016diversity} during inference to levitate the problem of generic responses (e.g., ``I don't know"). For all models, a simple re-rank operation is applied during inference to rank the generated responses $\hat{Y}$ based on their affective strength computed as $\frac{1}{|\hat{Y}|}\sum_{y \in \hat{Y}}||\overline{\textit{VAD}}(y)||_2$. All models are initialized with a uniform distribution in the $[-0.08, 0.08]$ interval, using the same seed. We trained all models with a batch size of 64 for 5 epochs using Adam \cite{kingma2014adam} optimization ($\beta_1 = 0.9$ and $\beta_2 = 0.999$) with the learning rate of 0.0001 throughout the training process.
 
\subsubsection{Results}
Table \ref{table: automatic} presents the results on model test perplexity in both MCT and PT (see Experiment 2). To analyze model generalization in different domains, we additionally report test perplexity on in-domain test dataset, which is created using 10K test pairs from the OpenSubtitles dataset. All models have comparable perplexity on both in-domain and out-domain test datasets, empirically showing that our proposed methods do not cause performance degradation in language fluency. One note is that the out-domain test perplexity for all models is quite large as compared to in-domain perplexity, as well as other dialog systems, e.g., \cite{vinyals2015neural}. One possible reason is that our testing dataset is different from the training dataset in terms of both vocabulary and linguistic distributions (the former was created from daily conversations, whereas the latter was created from movie subtitles). As a result, the models may not generalize well.
\begin{table}[!t]
\small
\centering
\begin{tabular}{p{1.5cm}||p{1.6cm}|p{1.1cm}|p{0.8cm}|p{0.8cm}}
 \hline
 \textbf{Experiment} & \textbf{Model} & \textbf{\#Params} & \textbf{PPL$\dagger$}& \textbf{PPL$\ddagger$} \\
 \hline
 \multirow{6}{*}{\parbox{1.5cm}{MCT (5M pairs)}}
 &S2S & 99M &42.5&124.3\\
 &S2S-UI& 99M &40.4&116.4\\
 &S2S-GI& 99M &40.7&120.3\\
 &S2S-LI& 99M &40.4&117.0\\
 &S2S-AO& 99M &40.2&115.7\\
 &AR-S2S& 99M &\textbf{39.8}&\textbf{113.7}\\
 \hline
 \multirow{2}{*}{\parbox{1.5cm}{PT (3M pairs)}}
 &S2S& 66M &41.2&130.6\\
 &S2S-Asghar& 66M &46.4&137.2\\
 &AR-S2S& 66M &\textbf{40.3}&\textbf{121.0}\\
 \hline
\end{tabular}
\caption{Model test perplexity. Symbol $\dagger$ indicates in-domain perplexity obtained on 10K test pairs from the OpenSubtitles dataset. Symbol $\ddagger$ indicates out-domain perplexity obtained on 10K test pairs from the DailyDialog dataset.}
\label{table: automatic}
\end{table}

Tables \ref{table: content} and \ref{table: emotion} present the evaluation results in MCT by five human annotators on the content quality and emotion quality, respectively. The values in brackets denote performance improvement in percentage. The Fleiss' kappa \cite{fleiss1973equivalence} for measuring inter-rater agreement is included as well. All models have ``moderate agreement'' or ``substantial agreement''. For content quality, all models except S2S-AO have noticeably more +2 ratings than S2S. For emotion quality, it is clear that all of our proposed affective models have significant improvement over S2S. Among the three affective attention mechanisms, S2S-LI achieves the best overall performance. Note that the improvement gained by affective attention and affective objective function are partially orthogonal. One explanation is that by actively paying attention to affect-rich words in the input sentence, our model is able to produce more accurate affect-rich words during decoding. Therefore, combing both techniques (AR-S2S) results in maximum improvement in emotion quality. Table~\ref{table: sample responses} presents some sample responses in the testing dataset.
\begin{table}[!t]
\small
\centering
\begin{tabular}{p{1.5cm}||p{0.4cm}|p{0.4cm}|p{0.4cm}|p{2.1cm}|p{0.8cm}}
 \hline
 \textbf{Model (\%)} & \textbf{+2} & \textbf{+1} & \textbf{0} &\textbf{Score}&\textbf{Kappa}\\
 \hline
 S2S   &22.4&47.0&30.6&0.918&0.544\\
 S2S-UI &\textbf{30.0}&48.6&21.4&\textbf{1.086 (+18.3\%)}&0.458\\
 S2S-GI &28.6&46.6&24.8&1.038 (+13.1\%)&0.413\\
 S2S-LI &29.4&47.2&23.4&1.060 (+15.5\%)&0.525\\
 S2S-AO &25.0&46.0&29.0&0.960 (+4.3\%)&0.482\\
 AR-S2S &29.6&44.8&25.6&1.040 (+13.3\%)&0.487\\
 \hline
\end{tabular}
\caption{Human evaluations on content quality (MCT).}
\label{table: content}
\end{table}
\begin{table}[!t]
\small
\centering
\begin{tabular}{p{1.5cm}||p{0.4cm}|p{0.4cm}|p{0.4cm}|p{2.1cm}|p{0.8cm}}
 \hline
 \textbf{Model (\%)} & \textbf{+2} & \textbf{+1} & \textbf{0} &\textbf{Score}&\textbf{Kappa}\\
 \hline
 S2S   & 19.0&33.2&47.8&0.712&0.613\\
 S2S-UI &23.6&36.0&40.4&0.832 (+16.9\%)&0.483\\
 S2S-GI &26.0&34.2&39.8&0.862 (+21.1\%)&0.652\\
 S2S-LI &24.6&36.4&39.0&0.856 (+20.2\%)&0.706\\
 S2S-AO &22.6&37.6&39.8&0.828 (+16.3\%)&0.602\\
 AR-S2S &\textbf{26.8}&37.2&36.0&\textbf{0.908 (+27.5\%)}&0.625\\
 \hline
\end{tabular}
\caption{Human evaluations on emotion quality (MCT).}
\label{table: emotion}
\end{table}
\begin{table}[!t]
\small
\centering
\begin{tabular}{p{2.1cm}||p{1.7cm}|p{3.1cm}}
 \hline
 \textbf{Message} & \textbf{Model} & \textbf{Response}  \\
 \hline
 \multirow{2}{*}{\parbox{2.1cm}{Mommy, can i stay up until eleven of the clock?}}
 &S2S (MCT) & Of course you can stay up late.\\
 \cline{2-3}
 &AR-S2S (MCT)& Of course you can, \textbf{sweetheart}. \\
 \hline
 \multirow{2}{*}{\parbox{2.2cm}{You are home late today, david. How was school?}}
 &S2S (MCT)& It was fine.\\
 \cline{2-3}
 &AR-S2S (MCT)& \textbf{Great fun} today.\\
 \hline
 \multirow{3}{*}{\parbox{2.1cm}{Do you like singing?}}
 &S2S (PT)& Yes, i do.\\
 \cline{2-3}
 &S2S-Asghar (PT)& I do not know.\\
 \cline{2-3}
 &AR-S2S (PT)& I \textbf{love} music.\\
 \hline
 \multirow{3}{*}{\parbox{2.1cm}{I'm pretty sure that jim will turn out to be a good lawyer.}}
 &S2S (PT)& He will turn out to be a good lawyer.\\
 \cline{2-3}
 &S2S-Asghar (PT)& I'm sure he will.\\
 \cline{2-3}
 &AR-S2S (PT)&The \textbf{best} lawyer in the world.\\
 \hline
\end{tabular}
\caption{Sample responses for models in both MCT and PT. Text in bold are affect-rich words.}
\label{table: sample responses}
\end{table}

\subsubsection{Analysis of Affective Attention}
\begin{figure}[!t]
\centering
\includegraphics[scale=0.30]{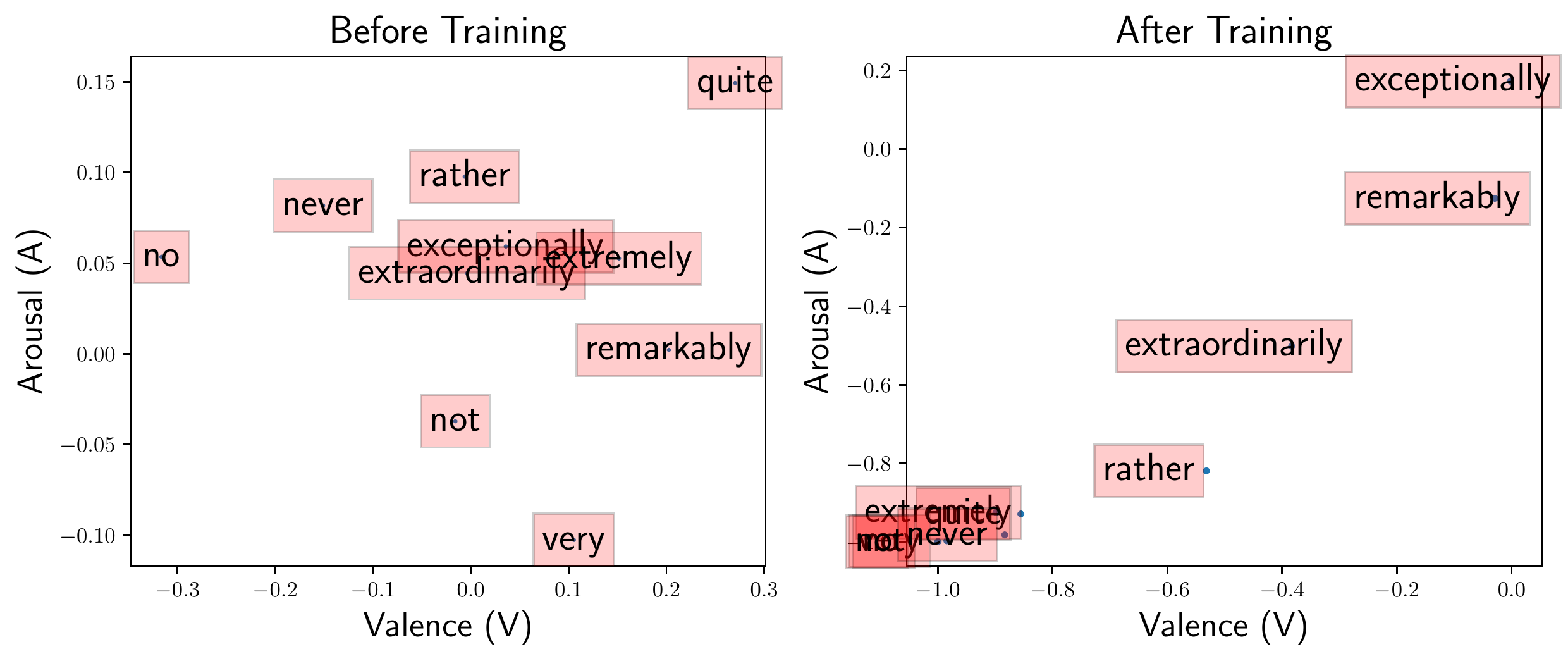}
\caption{Learned parameter $\beta$ (see equation (\ref{eqn: affective attention})) in Valence (V) and Arousal (A) dimensions for several common negators and intensifiers. Left sub-figure: before AR-S2S is trained. Right sub-figure: after AR-S2S is trained.}
\label{fig: bigram}
\end{figure}

To examine our hypothesis that our affective attention mechanism can correctly capture the effect of negators and intensifiers, we plot the learned parameter $\beta$ (see equation (\ref{eqn: affective attention})) in the Valence and Arousal dimensions in Figure \ref{fig: bigram}. It is obvious that our model successfully learned to make $\beta$ negative for negators. In addition, several extreme intensifiers such as ``exceptionally" and ``remarkably" have higher $\beta$ than less extreme intensifiers such as ``very'' and ``quite'', which is consistent with our hypothesis. One note is that our model does not learn well for some intensifiers such as ``extremely'', whose $\beta$ is comparable to less extreme intensifiers such as ``very". This result is not surprising because the impacts of intensifiers are difficult to be completely captured as they tend to vary depending on the following words \cite{kiritchenko2016the}. 

\begin{figure}[!t]
\centering
\includegraphics[scale=0.32]{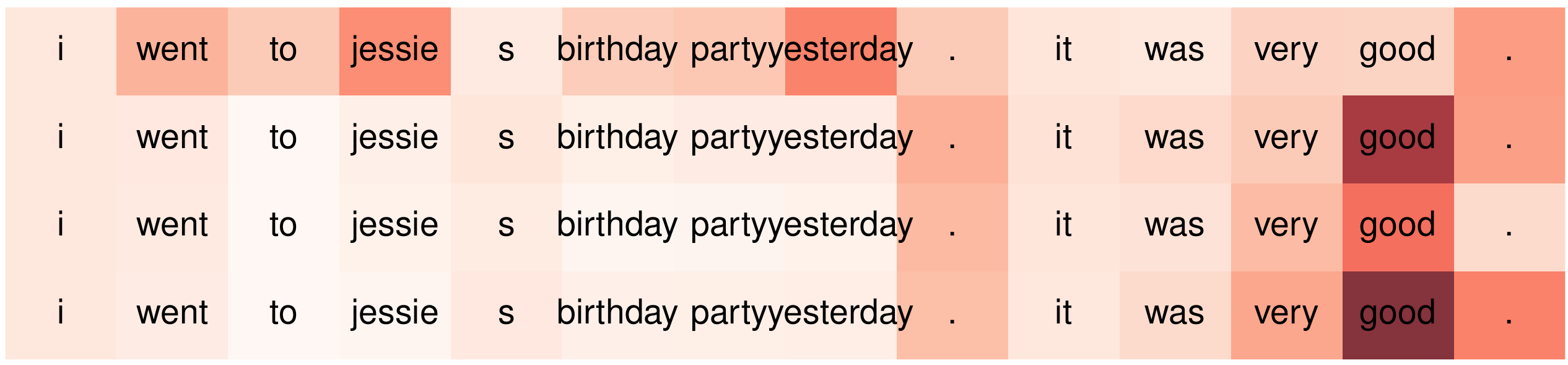}
\caption{Learned attention on a sample input sentence from the testing dataset. From top to bottom, the models are S2S, S2S-UI, S2S-GI and S2S-LI, respectively. Darker colors indicate larger strength.}
\label{fig: attn}
\end{figure}

Figure \ref{fig: attn} shows the attention strength over a sample input sentence in the testing dataset. As expected, our proposed affective attention models place extra attention on affect-rich words, i.e., ``good'' in this case. In addition, S2S-UI and S2S-LI have larger strengths than S2S-GI. This result is aligned with our model's assumption because different ``term importance'' have different impacts on the attention strengths and the word ``good'' here is quite common ($p(``good") = 0.00143$), which leads to the lower strength in S2S-GI.
\subsubsection{Analysis of Affective Objective Function}
We analyze the capability of our proposed affective objective function in producing affect-rich words. Table \ref{table: response stats} presents the number of distinct affect-rich words in randomly selected 1K test responses produced by different models. Affect-rich words are defined as words with VAD strength in $l_2$ norm exceeding the given threshold. It is clear that all S2S-AO models can produce more affect-rich words than S2S. In addition, the number of affect-rich words for every threshold increases steadily as the affective objective coefficient $\delta$ increases, showing a good controllability of our model via $\delta$.
\begin{table}[!t]
\small
\centering
\begin{tabular}{p{2.4cm}||p{1.3cm}|p{1.3cm}|p{1.4cm}}
 \cline{1-4}
  & \multicolumn{3}{|c}{\textbf{Threshold for $l_2$ Norm of VAD}}\\
 \hline
 \textbf{Model} & 3 & 2 & 1\\
 \hline
 S2S &25&104&190\\
 S2S-AO ($\delta=0.5$) &36&138&219\\
 S2S-AO ($\delta=1$) &50&154&234\\
 S2S-AO ($\delta=2$) &69&177&256\\
 \hline
\end{tabular}
\caption{Number of distinct affect-rich words (MCT).}
\label{table: response stats}
\end{table}
\subsection{Experiment 2: Preference Test (PT)}
\label{sec:PT}
\begin{table}[!t]
\small
\centering
\begin{tabular}{p{1.6cm}||p{1.1cm}|p{1.1cm}|p{1.1cm}}
 \cline{1-4}
  & \multicolumn{3}{|c}{\textbf{Threshold for $l_2$ Norm of VAD}}\\
 \hline
 \textbf{Model} & 3 & 2 & 1\\
 \hline
 S2S &21&83&157\\
 S2S-Asghar &31&120&217\\
 AR-S2S  &52&173&319\\
 \hline
\end{tabular}
\caption{Number of distinct affect-rich words (PT).}
\label{table: PT response stats}
\end{table}
We conduct human preference test to compare our \textbf{AR-S2S} with the state-of-the-art baseline \textbf{S2S-Asghar}, the best model proposed in \cite{asghar2017affective}. To the best of our knowledge, S2S-Asghar is the only model in the neural conversational model literature that aims to produce affect-rich responses in an end-to-end manner (i.e., without explicit user-input emotions). We also include \textbf{S2S} for comparison.

To make comparisons fair, we follow the specifications of S2S-Asghar and keep the number of parameters in all models comparable by reducing the size of our model. We use a smaller training dataset with 3 million random pairs and a vocabulary of size 20,000 due to the reduced model size. Note that our training dataset is still significantly larger than the original dataset used in \cite{asghar2017affective}, which comprises only 300K pairs and a vocabulary size of 12,000. All models have a word embedding of size 1027, a single-layer LSTM encoder and a single-layer LSTM decoder. All training specifications remain the same as the MCT except that S2S-Asghar is trained for 4 epochs with conventional cross-entropy loss and 1 more epoch with their proposed objective function, which includes a term to maximize affective content.

For human evaluation, we follow the same procedures as adopted in MCT except that five human annotators were asked to choose their preferred responses based on content quality and emotion quality, respectively, instead of annotating +2, +1 and 0. Ties are allowed.

\subsubsection{Results}
Table \ref{table: PT response stats} shows the number distinct of affect-rich words in randomly selected 1K responses produced by S2S, S2S-Asghar and our model. It is clear that our model produces significantly more affect-rich words than both S2S-Asghar and S2S. 

Table \ref{table: preference} shows the result of human evaluation. The Fleiss' kappa scores for content/emotion qualities are included in the last column. All models have ``moderate agreement'' or ``substantial agreement''. For content preference, our model scores relatively 21\% higher than S2S-Asghar. For emotion preference, our model scores relatively 50\% higher than S2S-Asghar. These findings show that our model is capable of producing better responses that are not only more appropriate in syntax and content, but also significantly more affect-rich than the state-of-the-art model.

\begin{table}[!t]
\small
\centering
\begin{tabular}{p{1.6cm}||p{1.7cm}|p{1.7cm}|p{1.4cm}}
 \hline
 \textbf{Model (\%)} & \textbf{Content} & \textbf{Emotion} & \textbf{Kappa}\\
 \hline
 S2S    &64 &26 &0.522/0.749\\
 S2S-Asghar &66 (+3.1\%)&32 (+23.1\%)&0.554/0.612\\
 AR-S2S    &\textbf{80 (+25.0\%)}&\textbf{49 (+88.5\%)}&0.619/0.704\\
 \hline
\end{tabular}
\caption{Human preference test (PT).}
\label{table: preference}
\end{table}
\subsection{Experiment 3: Sensitivity Analysis}
\label{sec:sensitivity analysis}
We examine the impacts of the affect embedding strength $\lambda$, affective attention hyper-parameter $\gamma$, as well as affective loss hyper-parameter $\delta$ on model perplexity and the number of affect-rich words produced. Due to the large number of experiments, we conduct the sensitivity analysis using 1 million pairs and a vocabulary of size 20,000. All training specifications remain the same as MCT except that the number of LSTM layers is 1, the hidden layer size is 512 and the embedding layer size is 303. 

\subsubsection{Results}
\begin{figure}[!t]
\centering
\includegraphics[scale=0.25]{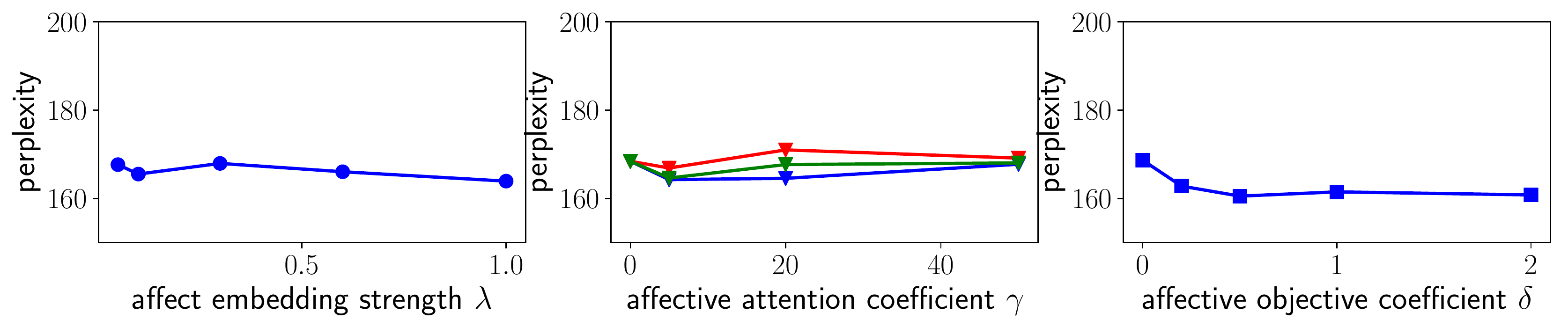}
\caption{Sensitivity analysis for affect embedding strength $\lambda$, affective attention coefficient $\gamma$, and affective objective coefficient $\delta$ on model perplexity. The blue, red and green curves (\textit{best viewed in color}) in the middle sub-figure represent $\mu_{ui}$, $\mu_{gi}$ and $\mu_{li}$ (see equation (\ref{eqn: term importance})), respectively.}
\label{fig: sensitivity}
\end{figure}

Figure \ref{fig: sensitivity} shows the plots of model test perplexity versus $\lambda$, $\gamma$ and $\delta$. Our model is fairly robust to a wide range of $\lambda$, $\gamma$ and $\delta$, regardless of the type of term importance. It is worth noting that the generated responses tend to become shorter with $\gamma \in [20, \infty]$, which may be caused by excessive attention placed on affect-rich words during decoding. Another interesting finding is that our affective objective function slightly improves test perplexity. One possible explanation is that affect-rich words are less common than generic words in our training corpus. As a result, our weighted cross-entropy loss placing extra weights on them improves the overall prediction performance.

\begin{figure}[!t]
\centering
\includegraphics[scale=0.25]{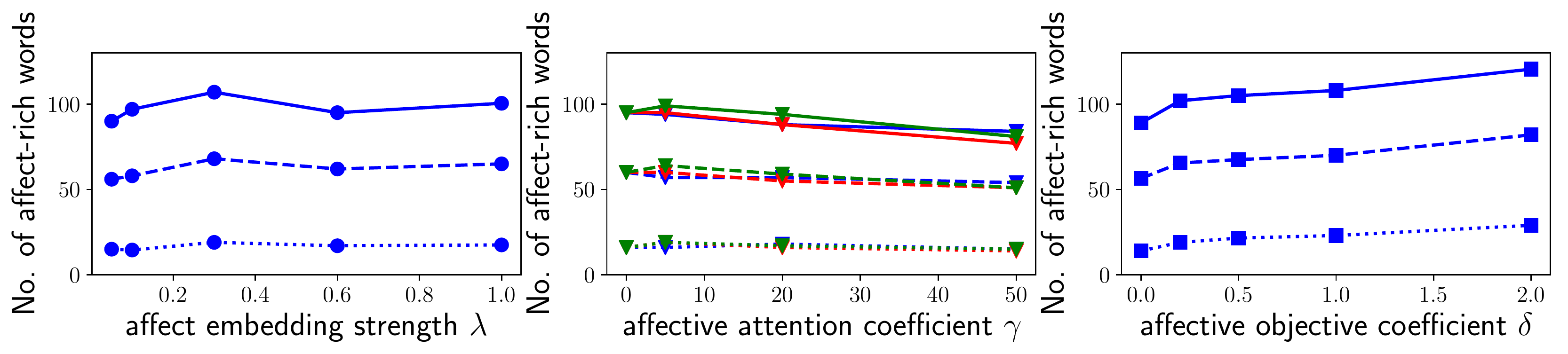}
\caption{Sensitivity analysis for affect embedding strength $\lambda$, affective attention coefficient $\gamma$, and affective objective coefficient $\delta$ on the number of distinct affect-rich words in randomly selected 1K test responses. The solid, dashed and dotted curves correspond to $l_2$ norm threshold of $1$, $2$ and $3$, respectively. The blue, red and green curves (\textit{best viewed in color}) in the middle sub-figure represent $\mu_{ui}$, $\mu_{gi}$ and $\mu_{li}$ (see equation (\ref{eqn: term importance})), respectively.}
\label{fig: sensitivity_words}
\end{figure}

Figure \ref{fig: sensitivity_words} shows the plots of the number of distinct affect-rich words in randomly selected 1K test responses versus $\lambda$, $\gamma$ and $\delta$. The number of distinct words increases slightly when $\lambda$ increases from 0 to 0.3, and then gradually decreases and stabilizes as $\lambda$ increases from 0.3 to 1. For $\gamma$ in all three term importance, there is an initial boost in the number of distinct words when $\gamma$ is small, i.e., $\gamma \in [0, 5]$. However, as $\gamma$ furher increases, the number of distinct words gradually decreases, which may be caused by limited word space during decoding due to excessive attention on affect-rich words. Among the three term importance proposed, local importance ($\mu_{li}$) is slightly more robust against $\gamma$ than the other two approaches. Finally, the number of distinct words consistently increases with $\delta$, which is similar to our findings from Table \ref{table: response stats}. Note that the numbers in this sensitivity analysis are much smaller than MCT, which can be attributed to smaller models and less training examples.

\section{Conclusion}
\label{sec:conclusion}
In this paper, we propose an end-to-end open-domain neural conversational model that produces affect-rich responses without performance degradation in language fluency. Our model leverages external word-VAD knowledge to encode affect information into the conversational model. In addition, our model captures user emotions by paying extra attention to affect-rich words in input sentences and considering the effect caused by negators and intensifiers. Lastly, our model is trained with an affect-incorporated weighted cross-entropy loss to encourage the generation of affect-rich words. Empirical studies on both model perplexity and human evaluations show that our model outperforms the state-of-the-art model of comparable size in producing natural and affect-rich responses.
\section{ Acknowledgments}
This research is supported, in part, by the National Research Foundation, Prime Minister's Office, Singapore under its IDM Futures Funding Initiative and the Singapore Ministry of Health under its National Innovation Challenge on Active and Confident Ageing (NIC Project No. MOH/NIC/COG04/2017 and MOH/NIC/HAIG03/2017).
\fontsize{9.5pt}{10.5pt}
\selectfont
\bibliography{aaai19}
\bibliographystyle{aaai}
\end{document}